\documentclass[runningheads]{llncs}

\usepackage{eccv}

\usepackage{eccvabbrv}

\usepackage{booktabs}
\usepackage{appendix}
\usepackage{graphicx}
\usepackage{algorithm, algpseudocodex, amssymb, array}
\usepackage{marvosym}
\usepackage[accsupp]{axessibility}  

\usepackage{hyperref}

\usepackage{orcidlink}

\begin{document}

\title{FreeDiff: Progressive Frequency Truncation for Image Editing with Diffusion Models} 

\titlerunning{FreeDiff}

\author{Wei Wu\inst{1,2} \and
Qingnan Fan\inst{2} \and
Shuai Qin\inst{2} \and
Hong Gu\inst{2} \and
Ruoyu Zhao\inst{3} \and\\
Antoni B. Chan\inst{1}\textsuperscript{(\Letter)}}

\renewcommand{\thefootnote}{}
\footnotetext[2]{\textsuperscript{(\Letter)} Corresponding authors}
\footnotetext[2]{Code is available at \href{https://github.com/Thermal-Dynamics/FreeDiff}{https://github.com/Thermal-Dynamics/FreeDiff}}

\authorrunning{W.~Wu et al.}


\institute{Department of Computer Science, City University of Hong Kong, Hong Kong, China \\
\email{weiwu56-c@my.cityu.edu.hk, abchan@cityu.edu.hk}\and
VIVO, Hangzhou, China \\
\email{fqnchina@gmail.com, \{shuai.qin,guhong\}@vivo.com} \and
Xidian University, Xi'An, China 
\email{royzhao@stu.xidian.edu.cn}}
\maketitle

\begin{abstract}
  Precise image editing with text-to-image models has attracted increasing interest due to their remarkable generative capabilities and user-friendly nature. However, such attempts face the pivotal challenge of misalignment between the intended precise editing target regions and the broader area impacted by the guidance in practice. Despite excellent methods leveraging attention mechanisms that have been developed to refine the editing guidance, these approaches necessitate modifications through complex network architecture and are limited to specific editing tasks. In this work, we re-examine the diffusion process and misalignment problem from a frequency perspective, 
  revealing that, due to the power law of natural images and the decaying noise schedule, the denoising network primarily recovers low-frequency image components during the earlier timesteps and thus brings excessive low-frequency signals for editing.
  Leveraging this insight, we introduce a novel fine-tuning free approach that employs progressive \textbf{Fre}qu\textbf{e}ncy truncation to refine the guidance of \textbf{Diff}usion models for universal editing tasks (\textbf{FreeDiff}). Our method achieves comparable results with state-of-the-art methods across a variety of editing tasks and on a diverse set of images, highlighting its potential as a versatile tool in image editing applications.
  \keywords{Diffusion Models \and Image Editing \and Frequency Truncation}
\end{abstract}
\section{Introduction}
\label{sec:intro}

In this work, we target the problem of text-driven image editing, which is a fundamental problem in computer vision and graphics. Although large-scale Text-to-Image (T2I) models have attracted increasing attention for multiple downstream vision tasks \cite{Noguchi_2024_WACV, Wang_2023_CVPR, Ceylan_2023_ICCV, hertz2022prompt, Cao_2023_ICCV, wu2023tune} due to their remarkable capacity for image generation and their user-friendly nature, leveraging T2I models for precise real-image editing tasks remains a significant challenge. As mentioned in several previous works\cite{hertz2022prompt, Cao_2023_ICCV}, while these models often succeed in introducing the specified elements to the image given the guidance from the text prompt (e.g., ``a hat''), they simultaneously induce unintended alterations in non-target areas, resulting in failed editing outcomes.

Recent approaches to using T2I models for image editing can be categorized into two paradigms. 
The first paradigm is based on fine-tuning of pre-trained T2I models based on a collection of text-image and image pairs to achieve an image-to-image (I2I) model \cite{nguyen2024visual,Han_2023_ICCV,Zhang_2023_CVPR}. It is labor-intensive and time-consuming and requires retraining according to the upgrade of the base T2I models, e.g. from SD v1.5 \cite{Rombach_2022_CVPR} to SDXL \cite{podell2023sdxl}. 
The second paradigm is a tuning-free approach, which purely relies on the feature manipulation of a pre-trained T2I model for image editing. Due to their convenience of deployment with only a pre-trained T2I model, a large number of works have emerged under this paradigm \cite{hertz2022prompt,Cao_2023_ICCV,Tumanyan_2023_CVPR}. 
The current state-of-the-art approaches in this paradigm use an inversion-reconstruction approach. 
First, inversion techniques\cite{song2020denoising,mokady2023null,pan2023effective} are applied to the image to recover the noisy latents that align with the model's prior distribution and that can accurately reconstruct the image contents. Next, editing methods are applied during the image re-generation process to refine the guidance encoded from the target prompt.
Previous editing methods (e.g., P2P\cite{hertz2022prompt}, PNP\cite{Tumanyan_2023_CVPR}, MasaCtrl\cite{Cao_2023_ICCV}) rely on manipulating the attention maps in the T2I model during the generation process, to reach a balance between preserving the fidelity of the non-target region and enabling editing capabilities.

However, a disadvantage of these attention manipulation methods is that they are highly specific to the image and the editing type (e.g., style, posture, identity replacement), and thus have limited \textit{versatility}, since different images require different hyperparameter settings, and limited \textit{generality}, since each manipulation only applies to one editing type. 
Thus, their complexity hinders the development of a unified approach that leverages their collective strengths simultaneously for universal editing and ease of use.

To address the aforementioned challenges, we propose \textit{FreeDiff}, a universal text-driven image editing approach, which is more compatible with various image editing types. 
Our approach is based on the following key observations: 
1) when using a text prompt to guide image editing, the generated editing effects are usually disrupted by various unwanted effects, while the desired editing effect only exists in the latent features of specific spatial frequency (SF) bands; 
2) During the denoising diffusion process for image generation, the image details are gradually increased in each step, demonstrating the gradual incorporation of higher frequency components into the image and latent space \cite{hertz2022prompt, Cao_2023_ICCV}.
3) Different image editing types require different levels of image details, for example, pose/shape edits correspond to low SF information, while identity replacement or texture changes correspond to high SF information.

Inspired by these observations and by examining the Fourier transform of the denoising network's intermediate features, we hypothesize that the network indeed prioritizes the learning of frequency components in a manner that correlates with the noise level across timesteps. Thus, to edit a specific image, proper guidance should mainly focus on specific frequency bands. Our analysis of the diffusion model justifies the common empirical findings and practices for different timesteps in image editing \cite{hertz2022prompt,Cao_2023_ICCV,Tumanyan_2023_CVPR}. 

Building on this analysis, we propose a novel fine-tuning free approach to image editing, which performs frequency truncation progressively to refine the guidance towards the target region. Initial hyperparameter settings are provided for different editing types, whereas better editing results can usually be obtained by fine-tuning the hyperparameters based on the initial settings. Empirical results from extensive image experiments demonstrate that our frequency space refinement of guidance facilitates versatile and universal editing capabilities.

The contributions of our work are summarized as follows: 
\begin{enumerate}
\item \textbf{Insights into the generation process from a spatial frequency perspective:} We provide a detailed analysis of a commonly observed phenomenon in the diffusion generation process, offering theoretical insights that lend an intuitive understanding of how the diffusion model’s learned prior conflicts with specific editing. 
\item \textbf{Innovation in Guidance Refinement:} We propose guidance refinement for real-image editing through spatial frequency techniques. This approach not only underscores the feasibility and versatility of SF-based methods in image editing but also introduces a novel alternative to attention map manipulation for guidance refinement.
\end{enumerate}

\begin{figure}[tb]
    \centering
    \includegraphics[width=0.88\textwidth]{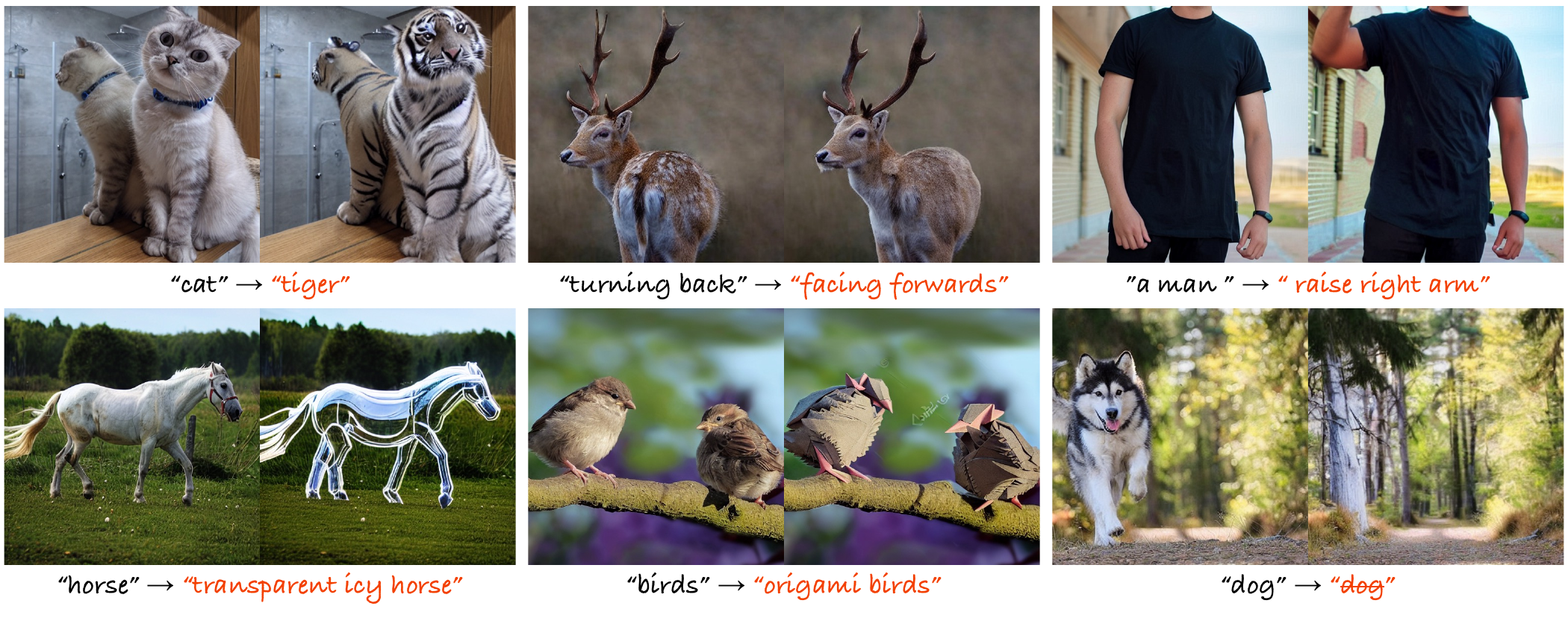}
    \captionsetup{width=\textwidth}
    \caption{Editing results across different editing tasks using our proposed method \textit{FreeDiff} demonstrate the effectiveness of our progressive frequency truncation strategy.}
    \label{fig:before_after}
\end{figure}

\section{Related Works}
\label{sec:related}
\subsection{Text-guided image editing}
Image editing with T2I diffusion models presents both an attractive opportunity and a formidable challenge due to the user-friendly nature of natural language input and the complex misalignment between guidance and the desired editing effects. Achieving a balance between fidelity in non-target regions and editing capabilities has been the focus of numerous studies. In contrast to relying on fine-tuning the T2I model\cite{Zhang_2023_CVPR,Han_2023_ICCV} for each specific image to be edited, fine-tuning free approaches\cite{hertz2022prompt,Cao_2023_ICCV,Tumanyan_2023_CVPR,meng2021sdedit} have gained popularity for their convenience. P2P\cite{hertz2022prompt} is the first to address the guidance misalignment issue through attention injection, specifically by swapping and re-weighting partial cross-attention maps between latent maps generated by the source and target prompts. PNP\cite{Tumanyan_2023_CVPR} examines the generation process and guides the editing by swapping specific self-attention maps at certain timesteps. MasaCtrl\cite{Cao_2023_ICCV} is proposed to tackle non-rigid editing tasks on which P2P and PNP fail (e.g., changing an object's pose), by substituting the query and key maps for certain layers and timesteps. While these methods have succeeded in specific editing tasks, they struggle with others; for example, MasaCtrl performs less satisfactorily in tasks involving changing an object to another or adding a new object. Furthermore, the various attention manipulation techniques proposed by these methods are difficult to unify into a general editing framework. 

In contrast to attention-based methods, our approach operates solely on the denoising network's output, without delving deep into the network structure, and is able to handle both rigid and non-rigid editing tasks.

\subsection{Inversion of diffusion}
Editing real images requires a tractable path through diffusion, making inversion techniques essential. DDIM inversion\cite{song2020denoising} with deterministic settings, employing a small guidance scale, is a simple and representative technique that leads to an acceptable image reconstruction with minor errors. However, the restriction imposed by the small guidance scale often conflicts with the requirements of many editing tasks. To address this issue, null-text inversion (NTI)\cite{mokady2023null} optimizes null embeddings for different diffusion timesteps to capture specific image information, leading to a better reconstruction result and overcoming the limitations of small guidance scales. Meanwhile, AIDI\cite{pan2023effective} introduces an accelerated fixed-point inversion method, enabling the application of larger guidance scales in subsequent editing tasks. EFI\cite{huberman2023edit} further enhances the editing capabilities by adding noise of different scales to the image to obtain noisy latents, which are then corrected with network inference, a process akin to virtual inversion. For simplicity, we use DDIM Inversion with fixed-point iteration to invert latents.

\section{Preliminaries}
\label{sec:prelim}
For simplicity, we provide a brief overview here while offering a detailed version in the Appendix Sec.A, which explains all symbols and provides additional details.

\noindent\textbf{Score-based diffusion models} The diffusion process can be implemented as different discretization formulations of Stochastic Differential Equation (SDEs)\cite{song2019generative, ho2020denoising, song2020score, song2020denoising }. To avoid the introduction of random noise during the inversion and generation processes and to make the analysis brief, in our study we adopt the deterministic DDIM formulation as in \cite{song2020denoising}. The marginal distribution of noise perturbed latent is:
\begin{align}
    p_{\sigma_t}(x_t|x_0) &= \sqrt{\alpha_t}\delta(x_0) +  \mathcal{N}(0, (1-\alpha_t)I), \quad \sigma_t^2 = 1 - \alpha_t, \quad t\in\{1,\cdots,T\},
    \label{eq:addnoise}
\end{align}
where $\delta(x_0)$ is a Dirac delta function centered at $x_0$ and $\alpha_t$ is the noise schedule coefficient. For brevity, we denote the score of the perturbed data as $\nabla_{x_t} \log p_{\sigma_t}(x_t)$.



\noindent\textbf{Guidance}
To control the generation, the guidance $g_t$ is commonly introduced by Classifier free guidance (CFG)\cite{ho2021classifier} as the difference between and conditional score\cite{dhariwal2021diffusion} $\epsilon_\theta(x_t, c)$ and unconditional score $\epsilon_\theta(x_t, \phi)$ as:
\begin{align}
    \nabla_{x_t} \log p_{\sigma_t}(c|x_t) &= \nabla_{x_t} \log p_{\sigma_t}(x_t|c) - \nabla_{x_t} \log p_{\sigma_t}(x_t) \\
    &= \epsilon_\theta(x_t, c) - \epsilon_\theta(x_t, \phi) = g_t,
    \label{eq:guidance}
\end{align}
The guidance $g_t$ is often enlarged by a factor $\gamma > 1$.



\noindent \textbf{DDIM Inversion}
Deterministic DDIM \cite{song2020denoising} inversion sample $x_t$ from $x_{t+1}$ by: 

\begin{equation}
    x_t = \frac{\sqrt{\alpha_{t}}}{\sqrt{\alpha_{t+1}}}x_{t+1} - \frac{\sqrt{\alpha_t(1-\alpha_{t+1})} - \sqrt{(1-\alpha_t)\alpha_{t+1}}}{\sqrt{\alpha_{t+1}}}\hat{\epsilon}_{\theta}(x_{t+1}).
    \label{eq:ddim}
\end{equation}
and $\hat{\epsilon}_{\theta}(x_{t+1})$ can be approximated by $\hat{\epsilon}_{\theta}(x_t)$:
\begin{equation}
    x_{t+1} = \frac{\sqrt{\alpha_{t+1}}}{\sqrt{\alpha_{t}}}x_t + \frac{\sqrt{\alpha_t(1-\alpha_{t+1})} - \sqrt{(1-\alpha_t)\alpha_{t+1}}}{\sqrt{\alpha_{t}}}\hat{\epsilon}_{\theta}(x_t).
\end{equation}



\section{Method}
In this section, we first give an analysis of the guidance provided by the denoising network during the diffusion process and illustrate how the network's learned prior conflicts with editing a specific image in Sec. \ref{sec:method-analysis}. Then, we detail the accordingly designed progressive truncation method in Sec. \ref{sec:method-freqtrunc}.
\label{sec:method}

\subsection{Diffusion prior from a frequency perspective}
\label{sec:method-analysis}
\begin{figure}[tb]
    \centering
    \includegraphics[width=0.85\textwidth]{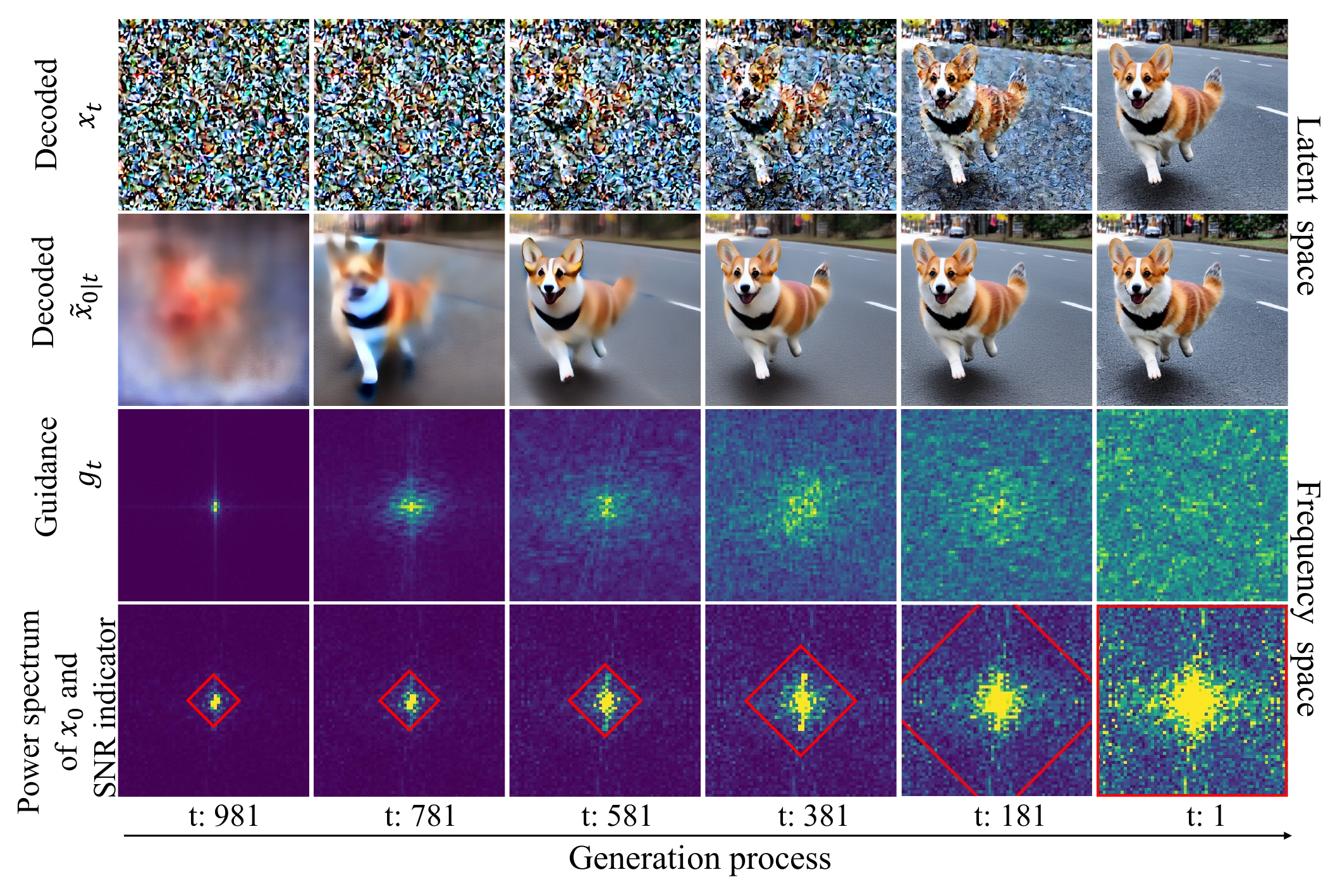}
    \captionsetup{width=\textwidth}
    \caption{Visualized decoded intermediate features and Fourier transformed features from a generation process with SD v1.5\cite{Rombach_2022_CVPR}, with the prompt ``a lovely corgi running on a city street''. The first, second, third, and fourth rows display the decoded noisy latents $x_t$s, the decoded $\tilde{x}_{t:0}$s, the guidance $g_t$, and the power spectrum of $x_0$ with the SNR (signal-to-noise ratio) indicator (red box) at the corresponding timestep. The timestep is shown at the bottom. The SNR box indicates where the signal (image) to latent noise ratio is greater than 1, which suggests the frequency bands that the network has higher probability to successfully recover $x_0$ from $x_t$.
    Note that to show lower frequency components, the same power spectrum is normalized with lower truncated upper bound as $t$ decreases.}
    \label{fig:interm_features}
\end{figure}

\noindent \textbf{Qualitative analysis} To gain a deeper understanding of the generation process, we visually inspect several intermediate results from the generation process across all timesteps. Since the latent $x_t$ from different timesteps obeys the marginal distribution in \eqref{eq:addnoise}, given an intermediate noisy latent $x_t$, we can obtain a corresponding image latent $\tilde{x}_{0|t}$ with the trained network as:

\begin{equation}
    \tilde{x}_{0|t} = \tfrac{1}{\sqrt{\alpha_{t}}}x_{t} - \tfrac{\sqrt{1 - \alpha_{t}}}{\sqrt{\alpha_{t}}}\hat{\epsilon}_{\theta}(x_{t})
\end{equation}
We then visualize in Fig.~\ref{fig:interm_features} the decoded noisy latent $x_t$, the corresponding decoded intermediate image $\tilde{x}_{0|t}$, and the guidance $g_t$ (Eq.~\ref{eq:guidance}) from the frequency space using the 2D discrete Fourier transform (DFT). 
From the example, we observe decreasing noise from the decoded $x_0$ (first row), as expected from the denoising process. However, when observing the intermediate images, a distinct pattern emerges from $\tilde{x}_{0|t}$ (second row), revealing a process where finer details are progressively added across timesteps, which means higher frequency components are gradually incorporated. The difference between $x_t$ and $\tilde{x}_{0|t}$ is consistent with the nature of both $x_0$ and $\tilde{x}_{0|t}$ being the weighted sum of the noisy latent $x_t$ and the semantically guided $\hat{\epsilon}_{\theta}(x_t)$, with $g_t$ having a larger weight in $\tilde{x}_{0|t}$. The frequency distribution of $g_t$ aligns with the visual transformations observed in $\tilde{x}_{0|t}$. Similar observations are obtained when inspecting other examples (see the Appendix Sec.B).

\noindent \textbf{Analysis in frequency space}
As pointed out in previous work by Field\cite{field1997visual}, examining the amplitude spectrum of a natural image reveals that it reaches a peak at low frequencies and decreases unevenly across all directions that frequency increases, following the power law $1/f^{\beta}$. Although the constant $\beta=1.1$ is not precisely defined, with the falloffs potentially being steeper or shallower, it is observed that most pictures exhibit the highest energy at the lowest frequencies. This law extends to images generated by diffusion models (which are predominantly trained on natural images), as evidenced by the power spectrum of such an image presented in the fourth row of Fig.~\ref{fig:interm_features}.

In the diffusion model training process, the noise introduced at timestep $t$ by \eqref{eq:addnoise} is additive white Gaussian noise (AWGN), characterized by uniform power across the frequency spectrum. Consider the image ultimately generated and displayed in Fig.~\ref{fig:interm_features}, serving as one of the training samples with an energy spectrum following the power law. AWGN, with a \emph{constant} energy spectrum of $\sigma_t^2 = 1 - \alpha $ across all frequencies, is added to the encoded latent $x_0$. This summation results in varying signal-to-noise ratios (SNRs) across different frequency bands, with higher frequencies possibly being obfuscated by noise. The denoising network is tasked with learning to predict the original $x_0$ from the perturbed $x_t$ as accurately as possible. Yet, confined by the SNRs, the denoising network must primarily recover the low-frequency components at earlier timesteps (when the noise power is large), and progressively higher frequency components as the power of the AWGN decreases. In the fourth row of Fig.~\ref{fig:interm_features}, the region with SNR $\geq 1$ is roughly outlined with a red box, serving as an indicator of which frequency bands the network could recover from the image at timestep $t$.

\noindent\textbf{Misalignment with editing a specific image} When it comes to editing a specific image, a significant conflict arises from the denoising network's inherent preference for low-frequency components. This preference is generally consistent between small enough steps\cite{song2020denoising}. In addition to the network's learned prior, the common weighting schedule\cite{song2020denoising} also amplifies the low-frequency components. For timestep $t$, the weight coefficient $w_{g_t}$ for $g_t$ in the final output $x_0$ is:

\begingroup
\begin{align}
    w_{g_t} & = -\gamma \tfrac{\sqrt{\alpha_{t-1}(1-\alpha_t)} - \sqrt{(1-\alpha_{t-1})\alpha_t}}{\sqrt{\alpha_t}} \times \tfrac{\sqrt{\alpha_{t-2}}}{\sqrt{\alpha_{t-1}}} \times \tfrac{\sqrt{\alpha_{t-3}}}{\sqrt{\alpha_{t-2}}} \times \cdots \tfrac{\sqrt{\alpha_{1}}}{\sqrt{\alpha_{2}}} \\
    &= -\gamma\sqrt{\alpha_1}(\sqrt{\tfrac{1}{\alpha_t}-1} - \sqrt{\tfrac{1}{\alpha_{t-1}}-1} ). 
\end{align}
\endgroup
This sequence generally decreases (except for the last few steps) during the generation process. For instance, in a typical 50-step DDIM generation process, the weights $w_{g_{981}}=1.25$, $w_{g_{681}}=0.23$,  $w_{g_{181}}=0.046$ demonstrate a decreasing trend. Consequently, this bias towards low frequencies contradicts the requirements of specific image editing tasks, where modifications of specific frequency bands are necessary.

\begin{figure}[tb]
    \centering
    \includegraphics[width=0.88\textwidth]{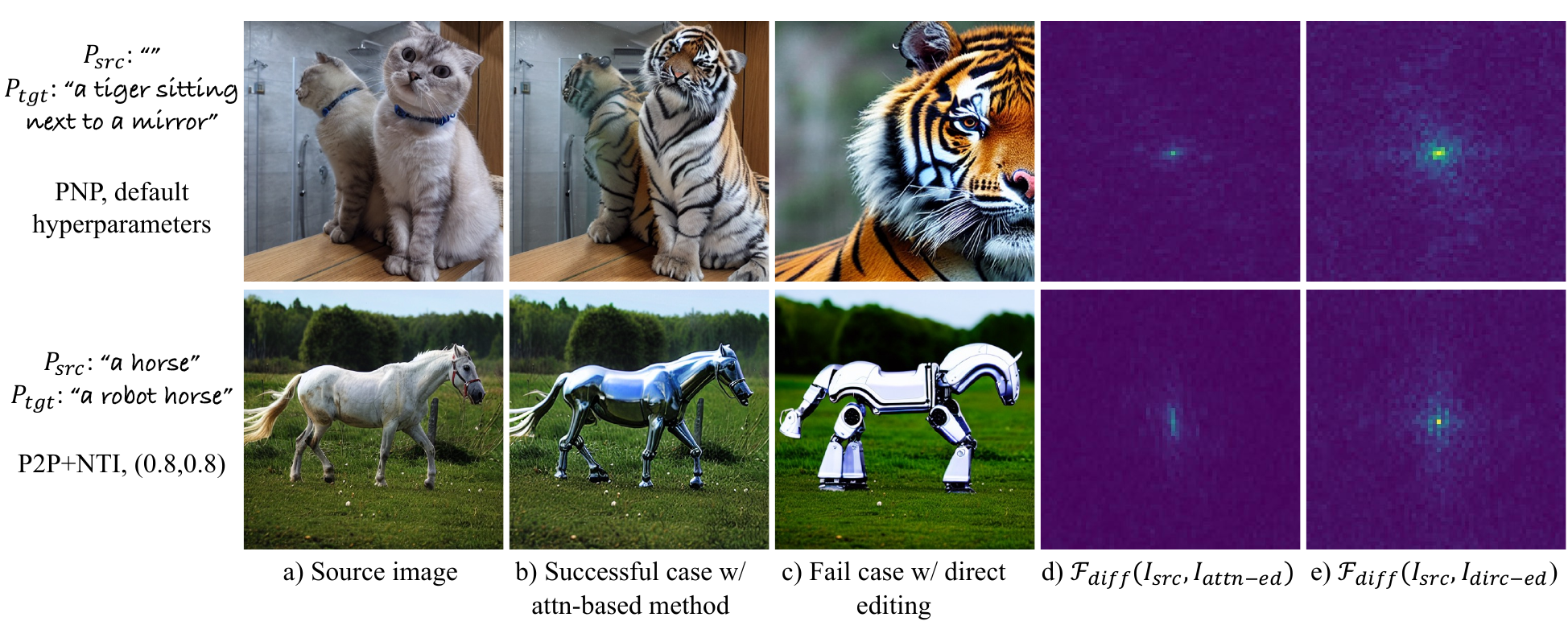}
    \captionsetup{width=\textwidth}
    \caption{Editing results from attention-based refining methods P2P\cite{hertz2022prompt}+NTI\cite{mokady2023null}, PNP\cite{Tumanyan_2023_CVPR}+fixed-point inversion in Section \ref{sec:method-freqtrunc} and directly applying guidance. Column d) and e) shows $\mathcal{F}_{diff}(I_{src}, I_{edit})$ between <source image, attention-based editing>, <source image, direct editing>, respectively. The $\mathcal{F}_{diff}(I_{src}, I_{edit})$ is normalized to the same numerical scale in each row. The results suggest that direct editing introduces low-frequency components with higher amplitudes.}
    \label{fig:fail-and-succ}
\end{figure}

Define the frequency difference between two encoded images, $I_1$ and $I_2$, as
\begin{align}
    & \mathcal{F}_{diff}(I_1, I_2) = \sum_{i=1}^C{\mathrm{abs}[\mathcal{F}\{x_{0:I_1}\}(\omega) - \mathcal{F}\{x_{0:I_2})\}(\omega)}],\\
    & x_{0:I_1} = \mathcal{E}(I_1), \quad x_{0:I_2} = \mathcal{E}(I_2)
\end{align}
where $\mathcal{E}$ denotes the encoder that transforms an image into its latent representation, 
and the summation performs channel-wise sum, with $C$ as the number of channels in the latent.
$\mathcal{F}\{x\}(\omega)$ is the 2D frequency transform (2D DFT) of image $x$, and $\omega$ is the spatial frequency variable.
In Fig.~\ref{fig:fail-and-succ} we visualize $\mathcal{F}_{diff}(I_{src}, I_{dirc-ed})$ and $\mathcal{F}_{diff} (I_{src}, I_{attn-ed})$ for the source image $I_{src}$ and edited images based on direct editing $I_{dirc-ed}$ and attention-based editing $I_{attn-ed}$, which supports our hypothesis that a successful editing introduces less power in low-frequency components. More examples of different editing types with different attention-based methods supporting the hypothesis are provided in the Appendix Sec.C.

\subsection{Progressive frequency truncation}
\label{sec:method-freqtrunc}
Based on the analysis and observations, we propose performing progressive truncation on guidance in the frequency space to achieve universal guidance refinement, allowing for both rigid and non-rigid editing within the same framework.

\noindent\textbf{Fixed-point DDIM Inversion} Similar to other fine-tuning free methods, our approach to editing begins by obtaining a suitable inverted latent $x_T$ from the encoded image $x_0$. Note that without approximation in \eqref{eq:ddim}, the inversion process is represented by an implicit function $x_{t+1} = f(x_{t+1})$:

\begin{equation}
    x_{t+1} = \tfrac{\sqrt{\alpha_{t+1}}}{\sqrt{\alpha_{t}}}x_t + \tfrac{\sqrt{\alpha_t(1-\alpha_{t+1})} - \sqrt{(1-\alpha_t)\alpha_{t+1}}}{\sqrt{\alpha_{t}}}\hat{\epsilon}_{\theta}(x_{t+1}), 
\end{equation}
which usually can be solved numerically through the iterations:
\begin{equation}
    x_{t+1}^{i+1} = f(x_{t+1}^i), \quad x_{t+1}^0 = f(x_t), \quad i\in\{0, \cdots, N\},
\end{equation}
where a small number of iterations, such as $N=3$ or $N=5$, on each inverting step, is sufficient to achieve nearly perfect reconstruction in most situations. It is worth noting that although fixed-point DDIM Inversion works in most cases, there is no guarantee of absolute correctness. If it fails to reconstruct the input image correctly, our method will be affected.

\begin{figure}[tb]
    \centering
    \includegraphics[width=0.9\textwidth]{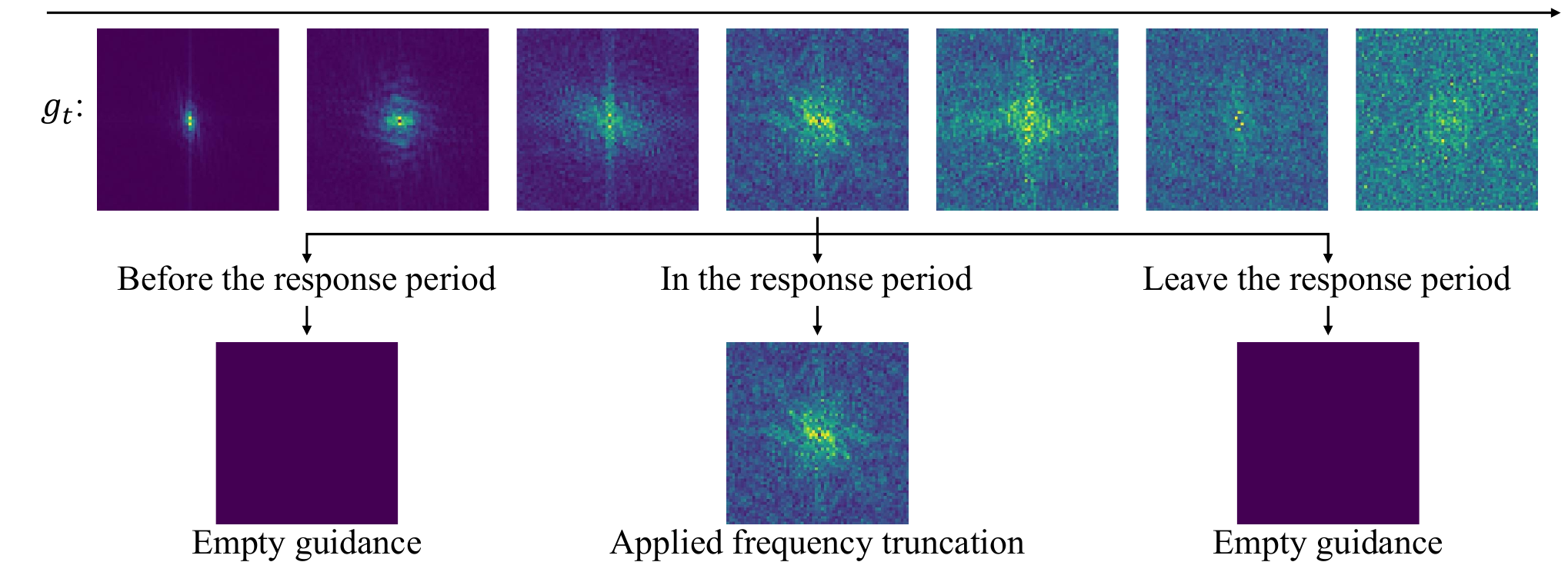}
    \captionsetup{width=\textwidth}
    \caption{The pipeline of our proposed method. The progressive frequency truncation is only applied in the response period according to Alg.~\ref{Alg:prog}, while guidance outside the response period is set to zero.}
    \label{fig:pipeline}
\end{figure}

\noindent\textbf{Frequency truncation with effective frequency band} Once we have the inverted latent $\hat{x}_T$, 
we apply frequency truncation to the guidance during the generation process. This approach is predicated on two key assumptions:
\begin{enumerate}
\item The generation process encompasses guidance through a single, continuous period associated with an atomic editing command (denoted as the response period), while the guidance outside this period is less relevant. An atomic editing command may involve changing, adding, or removing a single object, etc.
\item Throughout the response period, an Effective Frequency Band (EFB) is essential for introducing modifications accurately without extra alterations in non-target regions. 
\end{enumerate}

Assuming the current timestep $t$ is within the response period and the EFB is known at the current step, we can refine our guidance $g_t$ by:

\begin{align}
    &\hat{g}_t = \mathrm{IFFT}(\mathrm{FFT}(g_t)\circ\mathcal{M}_t^H(r)\circ\mathcal{M}_t^L(r)), \\
    &\mathcal{M}_t^H(r) = \mathcal{I}(r>r_t^H), \quad \mathcal{M}_t^L(r) = \mathcal{I}(r<r_t^L), 
\end{align}
where $\mathcal{I}$ is an indicator function that gives $1$ and $0$. $\mathcal{M}_t^H(r)$ and $\mathcal{M}_t^L(r)$ are high-pass and low-pass filters at timestep $t$ and $r_t^H$ and $r_t^L$ are the threshold radii for the corresponding 2D frequency filter. $\circ$ denotes element-wise multiplication. $\mathrm{FFT}(\cdot)$ and $\mathrm{IFFT}(\cdot)$ are the 2D Fourier transform and inverse 2D Fourier transform, respectively. In practice, we empirically select the response period and $r_t^H$s, $r_t^L$s according to each editing type (see the next section).

In addition to performing filtering on guidance in frequency space, we also zero out the guidance pixels whose alteration is above a threshold after inverse Fourier transform, since the power of these pixels mainly consists of low frequencies. And we further zero out the $80\%$ smallest values, most of which are already $0$s after the previous truncation:

\begin{align}
    & \mathcal{M}_t^S = \mathcal{I}\big(\frac{\mathrm{abs}(\hat{g}_t - g_t)}{\mathrm{abs}(g_t)} < \kappa \big), \quad \tilde{g}_t = \hat{g}_t \circ \mathcal{M}_t^S,\\
    & \mathcal{M}_t^V = \mathcal{I}\big(\tilde{g}_t > \eta_{0.8}(\tilde{g}_t)\big), \quad g_t^{\ast} = \tilde{g}_t \circ \mathcal{M}_t^V,
\end{align}
where $\kappa=0.6$ is the threshold, and $\eta_{0.8}(\tilde{g}_t)$ denotes the $80\%$ percentile of $\tilde{g}_t$. We term this as $\eta$ truncation, as opposed to the main spatial frequency truncation technique.
Given these assumptions, the implementation of the progressive frequency truncation algorithm is 
detailed in Alg.~\ref{Alg:prog} and visualized in Fig.~\ref{fig:pipeline}.

\begin{algorithm}[!htb]
\caption{Progressive Frequency truncation}\label{Alg:prog}
\textbf{Input:} Inverted $\hat{x}_T$, Start and end timestep of response period $T_{st}$, $T_{ed}$, Low-pass and \\ \hspace*{10mm}  high-pass filter pairs and their upperbound timestep $\{(r_t^H, r_t^L, \tau_i)\}$\\
\textbf{Output:} Refined guidance sequence $\{g_t^{\ast}\}$

\begin{algorithmic}[1]
\State $i = 1$
\For{$t = T, T-1, \dots, 1$}
    \If{$t>T_{st}$ \textbf{or} $t<T_{ed}$}
        \State $g_t=0$
        \State \textbf{continue}
    \Else
        \If{$t>=\tau_i$}
        \State $i = i+1$
        \EndIf
        \State $\mathcal{M}_t^H(r) = \mathcal{I}(r>r_t^H), 
        \quad \mathcal{M}_t^L(r) = \mathcal{I}(r<r_t^L)$
        \State $\hat{g}_t = \mathrm{IFFT}(\mathrm{FFT}
        (g_t)\circ\mathcal{M}_t^H(r)\circ\mathcal{M}_t^L(r))$
        \State $\mathcal{M}_t^S = \mathcal{I}(\frac{\mathrm{abs}(\hat{g}_t - g_t)}{\mathrm{abs}(g_t)} < 0.6 )$
        \State $\tilde{g}_t = \hat{g}_t \circ \mathcal{M}_t^S$
        \State $\mathcal{M}_t^V = \mathcal{I}(\tilde{g}_t > \eta_{0.8}(\tilde{g}_t))$
        \State $g_t^{\ast} = \tilde{g}_t \circ \mathcal{M}_t^V$
    \EndIf
\EndFor
\State \textbf{Return} $\{g^{\ast}_t\}$
\end{algorithmic}
\end{algorithm}

\subsection{Application to editing}
The categorization of editing types, when viewed from the perspective of frequency, varies significantly.
Identity replacement, such as transforming a dog into a lion, is akin to object removal, with both focusing on modifications of image textures, i.e., high spatial frequency (SF) information. In contrast, alterations in shape and pose correspond to adjustments in lower SF information. Changes in color and environmental (color) adjustments are associated with the lowest SF components, requiring specialized handling. 

Given that the guidance amplitude for color in diffusion models is relatively low compared to that for objects, and since color information typically aligns with the lowest frequency components, merely applying frequency truncation proves ineffective for color editing. Instead, we propose a \textit{two-step process} for color changes: first, generating a coarse mask for the object whose color is to be altered through frequency truncation on the guidance describing it at specific timesteps, for instance, "a white hat". Then, utilizing this mask solely to perform a guidance truncation for the edit. The approach for changing the environment surrounding an object follows a similar methodology.

Empirically, we define hyperpermeter sets $(T_{st}, T_{ed}, r_t^H, \tau_i)$ for each editing type. In practical applications, $r_t^L$ is rarely employed due to its minimal impact in most scenarios. Detailed values for these hyperparameters are provided in the Appendix Sec.D.

\begin{figure}[!htb]
    \centering
    \includegraphics[width=0.85\textwidth]{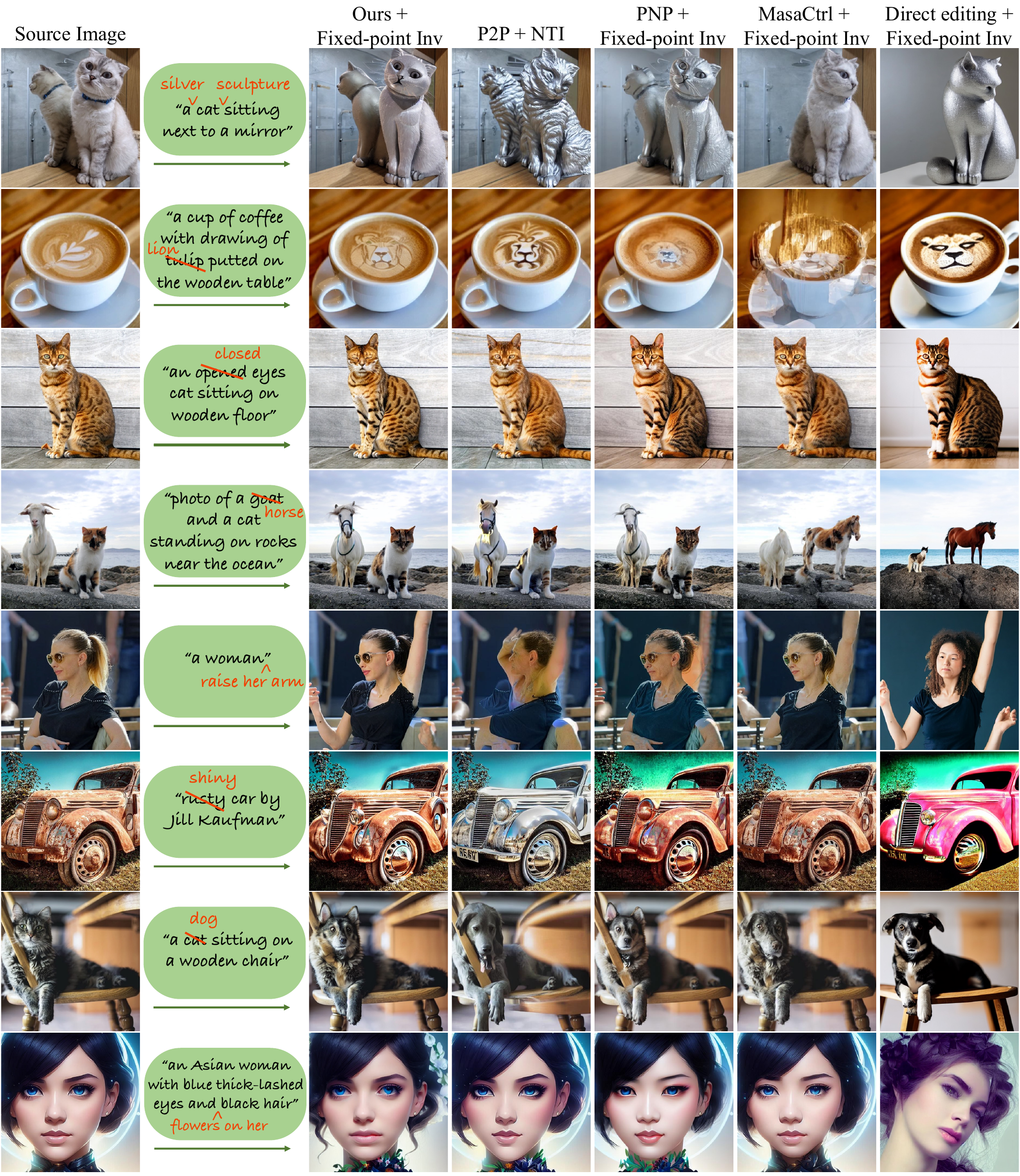}
    \captionsetup{width=\textwidth}
    \caption{Qualitative results comparing with 3 typical attention-based editing methods: P2P, PNP, MasaCtrl on images from the PIE dataset\cite{ju2023direct}. Direct editing results with fixed-point inversion are also shown as a baseline.}
    \label{fig:qual_pre}
\end{figure}

In summary, we propose a novel approach that refines editing guidance through progressive frequency truncation. Our method offers an effective guidance refinement strategy and default hyperparameters, with the fine-tuning of hyperparameters for a better editing result left to the users's aesthetic judgment,
consistent with P2P\cite{hertz2022prompt}, PNP\cite{Cao_2023_ICCV}, and MasaCtrl\cite{Tumanyan_2023_CVPR} that have adjustable editing hyperparameters. However, in contrast to previous methods that work well on certain editing types, our method can be used for a wider variety of editing types. 
 Our method's effectiveness is validated through successful edits across a diverse set of images and editing types (see Fig.~\ref{fig:before_after} and Fig.~\ref{fig:qual_pre}).

\section{Experiments}
\label{sec:exps}

 We evaluate our method on 
 on a variety of editing tasks and on a diverse set of images, most of which are sourced from the PIE benchmark\cite{ju2023direct}. For a fair comparison, all methods use the same SD v1.4 or SD v1.5\cite{Rombach_2022_CVPR} checkpoint, following the corresponding official implementations. Fixed-point iteration with $N=5$ is used for PNP, MasaCtrl and Our method to factor out the effect of inversion, except in cases where reconstruction fails with $N=5$. Null-text inversion is employed for P2P as it requires word-aligned source and target prompts to refine guidance with prompt-based attention operations. For inversion and re-generation with the target prompt, we perform the DDIM deterministic forward and backward sampling for 50 steps for all methods. The guidance scale is set to $7.5$ and the editing prompt remains the same across all methods unless specified otherwise. 

\subsection{Qualitative results}
We present typical qualitative results that demonstrate the comparative performance of our method against attention-based methods, including P2P, PNP, and MasaCtrl, as illustrated in Fig.~\ref{fig:qual_pre}. All showcased images are sourced from the PIE dataset. As evinced in Fig.~\ref{fig:qual_pre}, our method succeeds in performing precise editing across both rigid and non-rigid tasks, whereas P2P and PNP fail to handle non-rigid tasks, such as pose changes in row 3 and row 5. Our method achieves a satisfying balance between fulfilling the editing purpose and preserving the structure of the image. Moreover, our method is typically good at exerting subtle modifications, such as closing eyes and opening mouths, by leveraging our capability to adjust the SF truncation to suit the editing task.

\subsection{Quantitative results}

\begin{table}
\centering
\caption{Quantitative results from partial dataset of PIE\cite{ju2023direct}}
\label{table:quan}
\footnotesize
\begin{tabular}{c >{\centering\arraybackslash}p{1.5cm} >{\centering\arraybackslash}p{1.5cm} >{\centering\arraybackslash}p{1.5cm} >{\centering\arraybackslash}p{1.5cm}}
\toprule
 & Ours & P2P & PNP & MasaCtrl \\ \midrule
CLIP Score (whole image) $\uparrow$          & \textbf{25.5140}  & 24.7521  & 25.4717  & 24.6614\\
Background LPIPS $\downarrow$    &\textbf{11.14}   & 11.83 &  15.01 & 13.97\\ \bottomrule
\end{tabular}
\end{table}

We evaluate the CLIP score between the entire image and corresponding caption, and the LPIPS  of the background region on the PIE dataset to assess semantic consistency and background preservation, respectively. However, we have identified two main issues within the PIE dataset, including incorrect categorization of image-editing type pairs and an ill-defined category, which hinders evaluation accuracy. To mitigate these issues, we selected a subset of approximately 200 images from PIE for a comprehensive evaluation, with the result shown in Tab.~\ref{table:quan}. Additionally, to illustrate the editing effect across subcategories, quantitative results are provided in Tab.~\ref{table:rebuttle}. It is important to note that while the CLIP score serves as a metric for semantic consistency, it is not without limitations. Specifically, while the PNP method yields a higher CLIP score, its editing results are not comparable to those of the P2P method. For a detailed discussion of these issues and the rationale behind subset selection, please refer to the Appendix Sec.D.

\begin{table}[htbp]
\centering
\caption{Quantitative results for corrected semantic categories, 'Cat' denotes category and the number is consistent with the original PIE dataset.}
\label{table:rebuttle}
\footnotesize
\begin{tabular}{c|ccccc}
\hline
Methods\\(Clip Score$\uparrow$/LPIPS$\downarrow$) & Cat:1(n:77) & Cat:2(n:50) & Cat:3(n:27) & Cat:5(n:11) & Cat:7(n:38) \\ \hline
MasaCtrl& 24.57/.1661 & 24.83/.1001 & 25.58/.1810 & 26.92/.1043 & 25.01/.1190 \\ 
PNP     & 25.30/.1733 & 26.03/.1053 & 25.77/.1997 & 26.92/.1293 & 26.45/.1328 \\ 
P2P     & 24.78/.1340 & 25.11/.0889 & 24.02/.1768 & 27.14/.0835 & 25.76/.0941 \\ 
Ours    & 24.97/.1253 & 26.49/.0798 & 24.17/.1428 & 27.47/.1341 & 25.74/.0972 \\ \hline
\end{tabular}
\label{tab}
\end{table}

\begin{figure}[!htb]
    \centering
    \includegraphics[width=0.9\textwidth]{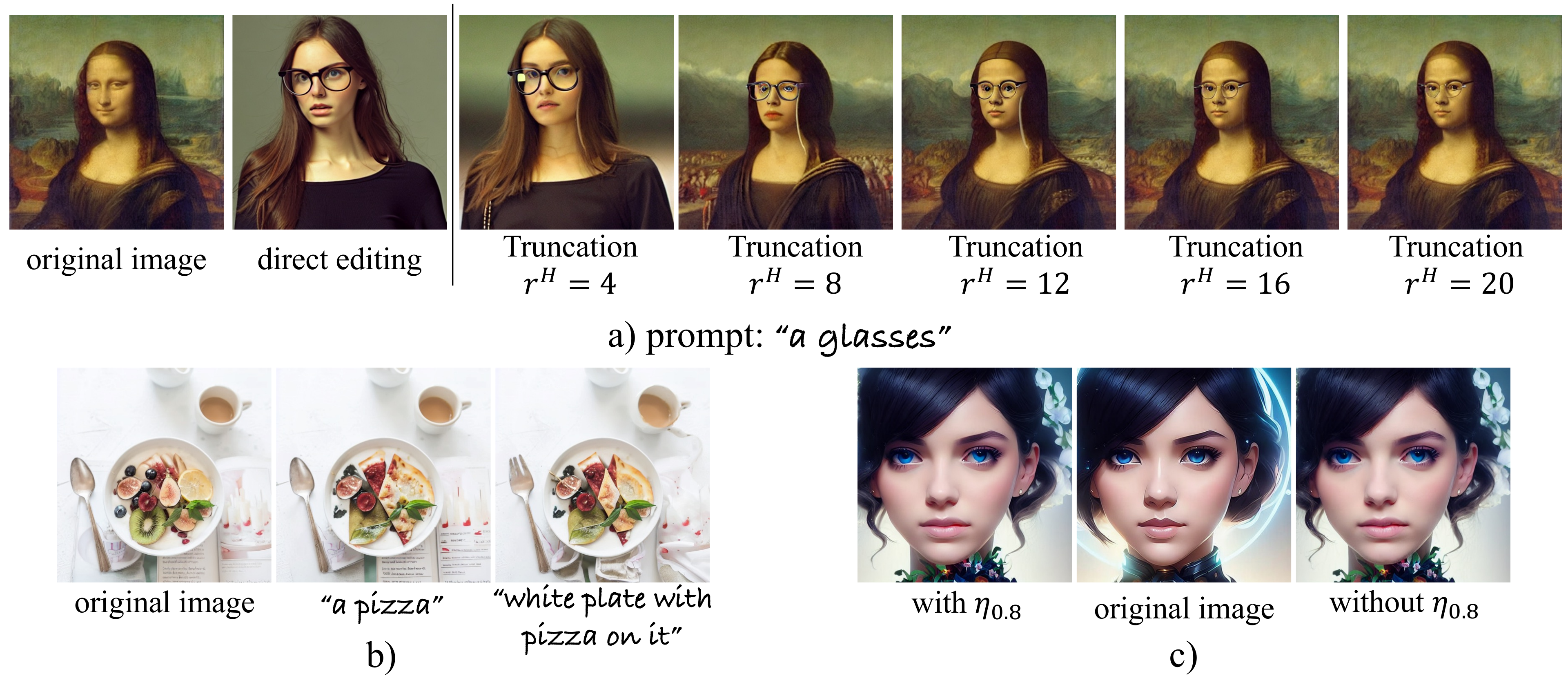}
    \captionsetup{width=\textwidth}
    \caption{a) shows the editing results with consistent truncating $r_t^H \in \{0,4,8,12,16,20\} x$ for all $t$, 
    and a prompt "a glasses". 
    The woman gradually becomes more akin to Mona Lisa and the ``glasses'' become less significant. b) shows the editing results with the same progressive frequency truncation hyperparameter sets but with different editing prompts. Our method prefers editing prompt that describes less editing-irrelevant content. c) shows how the editing results subtly changed when using $\eta$ truncation, which helps to preserve details. Note the difference in face shape and light on the hair.}
    \label{fig:ablation}
\end{figure}

\subsection {Ablation study}
\noindent\textbf{Effect of SF truncation without FreeDiff} 
Applying SF truncation directly throughout the whole generation process without FreeDiff yields outcomes that corroborate our hypothesis. For instance, by applying SF truncation within a smaller range for the first generation and larger ones for other generation for the same editing prompt, we can see in Fig.~\ref{fig:ablation}a that progressively enlarging the SF truncation range causes the edited image to more closely resemble the source image, while causing the intended edits (the glasses) gradually become less significant. This observation supports our assumptions, as detailed in Sec.~\ref{sec:method-freqtrunc}, regarding the existence of specific response periods and effective frequency bands.

\noindent\textbf{Sensitivity to editing prompt} FreeDiff, by applying SF truncation on the guidance, is highly affected by the guidance text. This effect is demonstrated in Fig.~\ref{fig:ablation}b, where under the same goal to turn the fruits on the plate into a pizza, a simple target prompt ``a pizza'' results in less background alteration compared with the PIE-provided target prompt "white plate with pizza on it". Generally, for editing we should avoid describing the objects and regions unrelated to the editing target.

\noindent\textbf{Effect of zero-out $\eta_{0.8}$ values} We demonstrate the effect of implementing $\eta$ truncation within FreeDiff. While $\eta$ truncation is not the primary mechanism driving precise editing outcomes, it helps with preserving the structure of non-editing regions. This subtle preservation effect is demonstrated in Fig.~\ref{fig:ablation}c, where using $\eta$ truncation preserves the light reflected on the girl's hair,
as well as the shape of her face.

\section{Limitations and Discussion}
In addition to being affected by erroneous reconstructions, our method is also constrained by the SD model's prior. Similar to MasaCtrl\cite{Cao_2023_ICCV}, our editing fails if the denoising network fails to generate a desired layout when changing an object's pose or color, or exactly locating on one of the multiple objects of the same kind. Our method is also sensitive to the description of the target prompt -- generally, a description with full contents that contain non-editing targets/regions hinders the structure preservation of our method, as demonstrated in the ablation study. 
To further improve our method, we will try to apply our \textit{two-step methods} to other editing types for a better result and combine our methods with attention manipulation techniques for a better control ability on image editing.

\section{Conclusion}
To the best of our knowledge, we are the first to explore frequency truncation with diffusion models for image editing by proposing FreeDiff, a novel tuning-free guidance refinement method without using attention-based manipulations. Our investigations reveal that applying guidance directly from a denoising network for editing a specific image leads to an unsatisfying result, primarily because the denoising network's learned prior tends to introduce excessive low-frequency components and affect the non-target regions. However, with the implementation of sophisticated spatial frequency truncation techniques, we demonstrate that it is entirely feasible to achieve precise editing with the guidance. FreeDiff does not depend on complex attention map manipulations and successfully tackles both rigid and non-rigid editing tasks within the same framework, marking a significant step towards a versatile and unified editing solution. 


\section*{Acknowledgements}
This work was supported by a Strategic Research Grant from City University of Hong Kong (Project No.7005840).

%
%
\bibliographystyle{splncs04}
\bibliography{main}

\newpage
\begin{appendix}
\section*{\centering\Large Appendix of "FreeDiff: Progressive Frequency Truncation for Image Editing with Diffusion Models"}

\end{appendix}
\setcounter{equation}{16}
\section{Preliminaries}
\textbf{Score-based diffusion models} The diffusion process, characterized by multi-level noise perturbations, can be formulated as the discretization of Stochastic Differential Equation (SDEs) \cite{song2020score} and can be reversed if the scores of all noise levels are known. Different discretization formulations lead to different diffusion models\cite{song2019generative, ho2020denoising, song2020score, song2020denoising}. Denote the encoded image latent as $x_0 \in \mathbb{R}^{CHW}$. The objective of the denoising network $\epsilon_\theta$ is to learn the score $\nabla_{x_t} \log p_{\sigma_t}(x_t|x_0)$ for the perturbed data $x_t$ across all noise levels $\sigma_t$ in the time step $t$\cite{song2019generative, ho2020denoising}:

\begin{align}
    \mathcal{L} &= \mathbb{E}_t[w(t)\mathbb{E}_{x_0}\mathbb{E}_{x_t|x_0}[\| \epsilon_{\theta}(x_t) - \nabla_{x_t} \log p_{\sigma_t}(x_t|x_0) \|_2^2]  ]
    \label{eq:diffusionloss} 
\end{align}
where $w(t)$ is a positive weighting function, $\alpha_t\in(0,1]$ is the noise schedule coefficient that controls the noise level and decreases to nearly $0$ as $t$ approaches $T$. 

\noindent\textbf{Guidance} To influence the generation process via conditional distributions, we focus on $\nabla_{x_t} \log p_{\sigma_t}(x_t|c)$, where the condition $c$ is the encoded embedding of the class labels, text prompt, etc. The conditional score \cite{dhariwal2021diffusion} can be expressed as:
\begin{equation}
    \nabla_{x_t} \log p_{\sigma_t}(x_t|c) = \nabla_{x_t} \log p_{\sigma_t}(x_t) + \nabla_{x_t} \log p_{\sigma_t}(c|x_t)
\end{equation}
Classifier free guidance \cite{ho2021classifier} is often used in T2I diffusion models as in Eq.2 and Eq.3, where $\epsilon_\theta(x_t, c)$ is the conditional score w.r.t.~the encoded text prompt $c$,  $\phi$ is the encoded embedding from a null (empty) string and $\epsilon_\theta(x_t, \phi)$ is its corresponding unconditional score. It is common practice to enlarge the guidance by a scaling factor $\gamma>1$ since $p^{\gamma}(c|x_t) \propto p(x_t|c)/p(x_t)$, which equals to enhancing the posterior probability.

\section{More intermediate features from generation process}

\begin{figure}[ht!]
    \centering
    \includegraphics[width=0.92\textwidth]{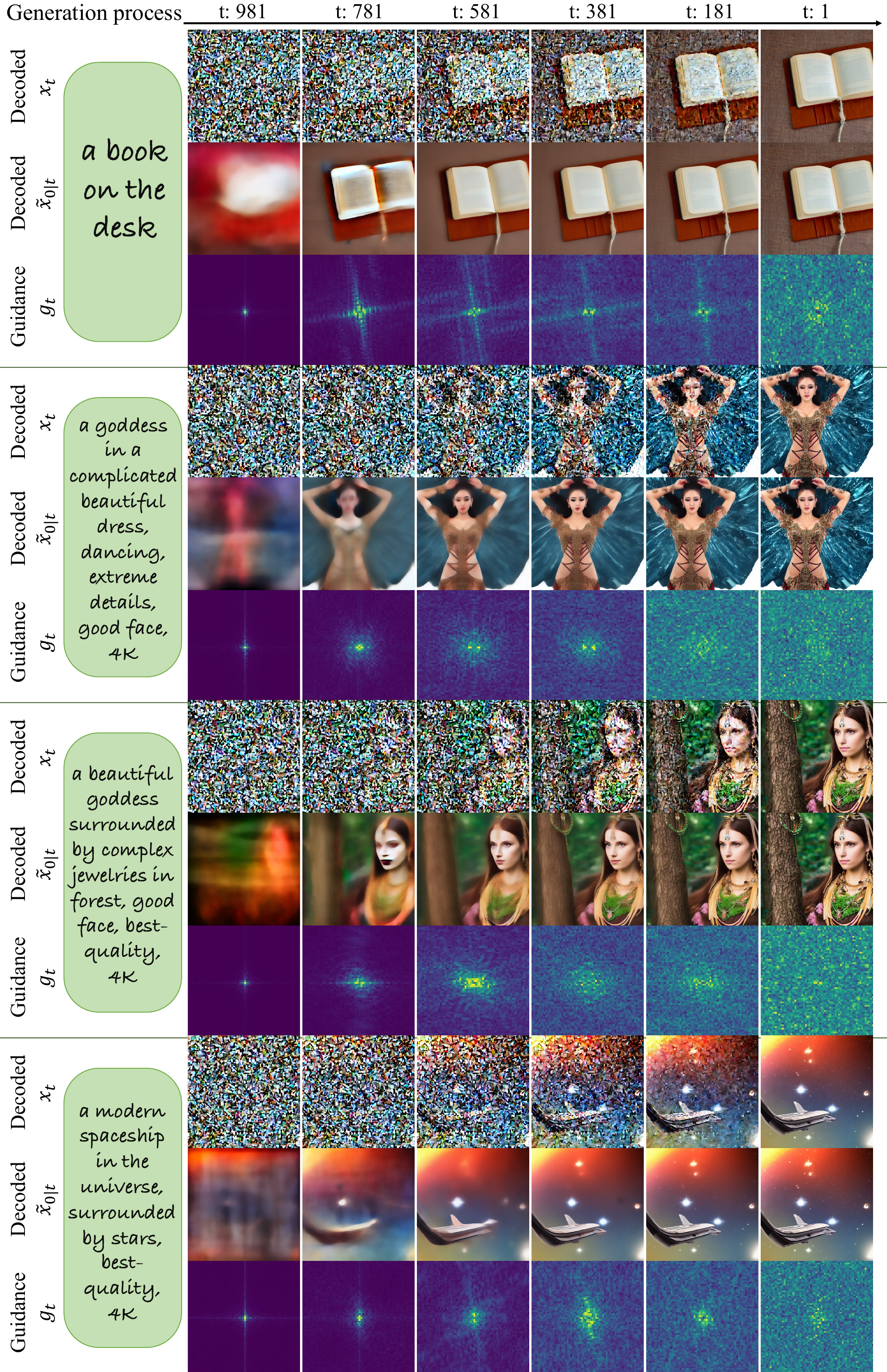}
    \caption{Visualized decoded intermediate features and Fourier transformed features from a generation process with SD v1.5 \cite{Rombach_2022_CVPR}.}
    \label{fig:gen1}
\end{figure}

\begin{figure}[ht!]
    \centering
    \includegraphics[width=0.92\textwidth]{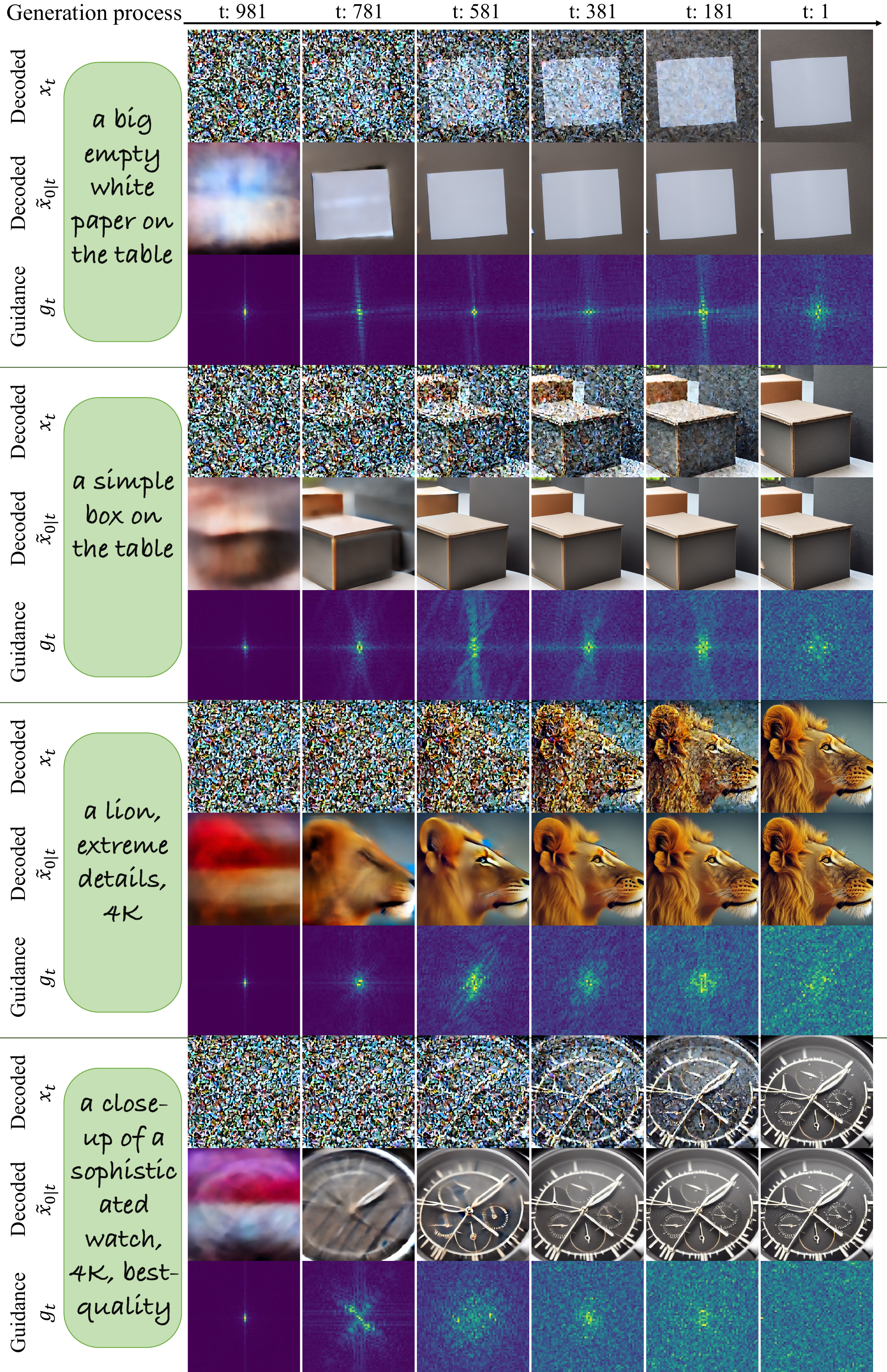}
    \caption{Visualized decoded intermediate features and Fourier transformed features from a generation process with SD v1.5\cite{Rombach_2022_CVPR}.}
    \label{fig:gen2}
\end{figure}

\label{sec:genexp}
More examples supporting our observations in the analysis section are listed in Fig.~\ref{fig:gen1} and ~\ref{fig:gen2}. The intermediate features are listed together with the prompt that generates the image. While these generated images show visual complexity of various levels, they follow a consistent generative pattern: details in $\tilde{x}_{0|t}$ are gradually added through the steps of generation, aligning with the gradual incorporation of higher frequency components from guidance.

\section{Editing difference from the frequency perspective}
\begin{figure}[ht!]
    \centering
    \includegraphics[width=\textwidth]{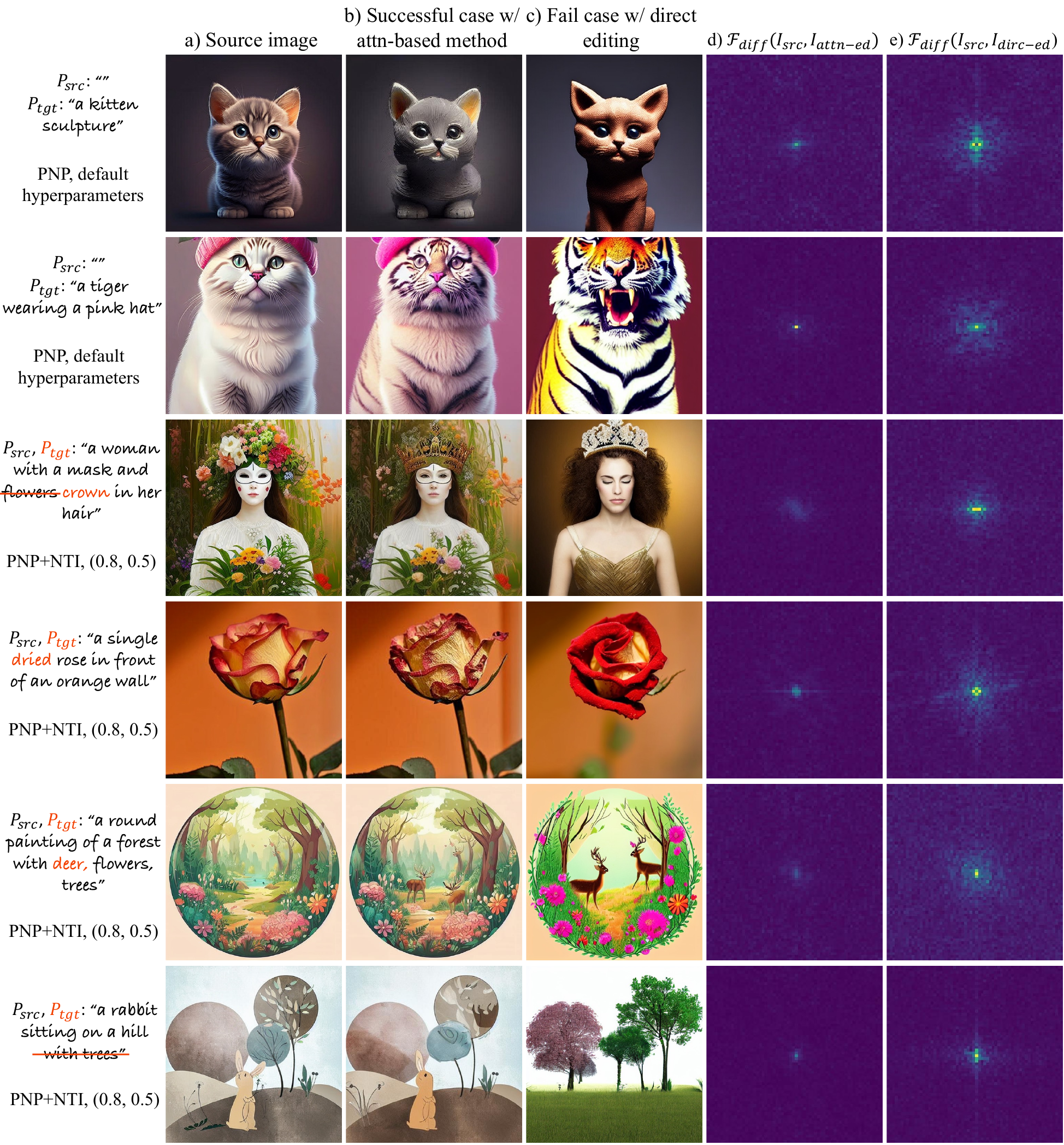}
    \caption{Editing results from ABMs: P2P \cite{hertz2022prompt}
    +NTI\cite{mokady2023null}
    , PNP\cite{Tumanyan_2023_CVPR}
    +fixed-point inversion and directly applying guidance. Column d) and e) shows $\mathcal{F}_{diff}(I_{src}, I_{edit})$ between <source image, attention-based editing>, <source image, direct editing>, respectively. The $\mathcal{F}_{diff}(I_{src}, I_{edit})$ is normalized to the same numerical scale in each row.}
    \label{fig:fail-and-succ-supp}
\end{figure}

Examples of various editing types applied to different images using two ABMs, P2P\cite{hertz2022prompt} 
and PNP\cite{Tumanyan_2023_CVPR}
, and direct editing are shown in Fig.~\ref{fig:fail-and-succ-supp}. These examples support our hypothesis that direct editing inadvertently introduces an excess of low-frequency components, due to the learned prior and weighting schedule of the denoising network, leading to an undesired alteration in non-target regions.

\section{Qualitative and quantitative results}
\label{sec:qualexp}
In this section, we first point out the existing problems within the PIE dataset [10]
, and then detail our categorization of editing types from the frequency perspective and provide a default hyperparameters set for reference. Both quantitative results and qualitative examples are listed.
\subsection{PIE Dataset}
The PIE\cite{ju2023direct}
dataset is the first large-scale dataset containing 700 images for quantitatively evaluating editing results across different editing types, with masks of target objects or regions provided for background-foreground assessment. However, there are two significant problems within the PIE dataset:

\begin{enumerate}
\item \textbf{Incorrect categorization} Within each editing type, some text-image pairs are misclassified. For example, in the "change object" category that aims at changing the identities of objects, the image "112000000008.jpg" is with prompt-pair "a painting of two women walking on the beach"-"a painting of two women walking on the grass", which would more aptly fit the "change background" category. Multiple misclassified editing pairs can be found in each category, hampering the dataset's credibility.
\item \textbf{Ill-defined category} In addition to the misclassification problem, the "change content" category, which primarily contains changes to objects, poses, materials, styles, encompasses only a minor portion of changes to shapes and expressions. These latter changes are more appropriate for the editing type "change content" to be distinguished from other types.
\end{enumerate}

These two problems hinder the accuracy of evaluation, since for ABMs, the best default hyperparameter sets vary largely for different editing types. For our method, accurately defining the editing types is crucial for selecting appropriate reference hyperparamters. Consequently, we will re-categorize the PIE dataset in the future, which will be detailed in the next section.

\subsection{Editing categories and hyperparameters with \textit{FreeDiff}}
With \textit{FreeDiff}, editing types are divided into three main categories from the frequency perspective:

\begin{enumerate}
\item \textit{SF-0}: Changes that primarily rely on low-frequency components. This category includes editing types such as changing colors, environments, poses, shapes, and adding objects with significant structural differences compared to the original region. These changes require the alteration of low-frequency components and are affected during the earliest generation steps. In the situations of changing colors and environments, a two-step method is required to refine the guidance instead of directly applying frequency truncation.
\item \textit{SF-1}: Changes that depend less on low-frequency components. Editing types in this category contain swapping object identities, removing an object, altering an object's material, changing style of the image, and adding objects. These changes rely less on low-frequency components and the editing mainly affects generation steps beyond the earliest.
\item \textit{SF-2}: Changes that solely rely on high-frequency components. Editing types in this category are similar to the second type but focus on small objects as targets. These changes only rely on high-frequency components in later generation steps.
\end{enumerate}

Given that our categorization primarily differentiates based on the spatial-frequency (SF) components involved, we denote these three categories as \textit{SF-0}, \textit{SF-1} and \textit{SF-2}, respectively, for brevity consideration.

For notation simplicity, we consolidate $T_{st}, T_{ed}$, and $\tau_i$ by setting $r_t^H$ to 32 outside the response period. With $r_t^H$ set to 32 and given a guidance map of 64x64 dimensions, the high-pass filter will block all the signals and zero-out the guidance. For \textit{SF-1}, one of the representative hyperparameter sets is $\{\tau_i = (781, 581, 1), r_t^H=(32, 10, 10)\}$, which means that we apply a high-pass filter with radii of 32, 10, and 10 for the time intervals $[981, 781]$, $(781, 581]$, and $(581, 1]$, respectively. As listed in Tab.~\ref{table:hyperparamers}, typical hypeparameter sets for \textit{SF-1} are $\{\tau_i = (781, 581, 1), r_t^H=(32, 10, 10)\}$, $\{\tau_i = (781, 581, 1), r_t^H=(32, 32, 10)\}$, $\{\tau_i = (681, 581, 481, 1), r_t^H=(32, 20, 8, 1)\}$. For \textit{SF-2}, typical hyperparameter sets are $\{\tau_i = (781, 581, 1), r_t^H=(32, 32, 20)\}$, $\{\tau_i = (781, 481, 1), r_t^H=(32, 32, 24)\}$. Notably, there are no typical hyperparameter sets for \textit{SF-0}.

\begin{table}
\centering
\caption{Hyperparameter sets for \textit{SF-0}, \textit{SF-1} and \textit{SF-2}}
\label{table:hyperparamers}
\begin{tabular}{c>{\centering\arraybackslash}p{3cm}>{\centering\arraybackslash}p{3cm}}
\toprule
SF Category & \multicolumn{2}{c}{Hyperparameters} \\ 
\cmidrule(lr){2-3} 
 & $\tau_i$ & $r_t^H$ \\ \midrule
SF-0 & N/A & N/A \\
SF-1 & (781, 581, 1) & (32, 10, 10) \\
     & (781, 581, 1) & (32, 32, 10) \\
     & (681, 581, 481, 1) & (32, 20, 8, 1) \\
SF-2 & (781, 581, 1) & (32, 32, 20) \\
     & (781, 481, 1) & (32, 32, 24) \\ \bottomrule
\end{tabular}
\end{table}

The choice of hyperparameter sets should primarily be based on the size of the object to be edited. We recommend using smaller high-pass filters in the earlier steps for editing larger objects.

\subsection{\textit{Two-step process} for editing colors and environments}
To change colors and environments, we apply a \textit{two-step process}. First, given that guidance truncated by \textit{FreeDiff} at each step typically has smaller values on each pixel and a higher ratio of pixels that are activated within the target region, we aggregate the truncated guidance maps across all timesteps to form a coarse mask for the target region. Then, we generate the target image by amplifying the guidance with this coarse mask, enhancing the refinement of the guidance. To preserve objects while changing the environment, the coarse mask can be reversed by subtracting it from a mask of ones. Some example results from this \textit{Two-step process} are demonstrated in Fig.~\ref{fig:qual-1}.

\subsection{Quantitative results}
For the overall quantitative results on the partial PIE dataset shown in Tab.1, we selected editing types where the comparison attention-based methods(ABMs) tended to perform well (change, add, and delete objects, change materials and poses). We did not include image where the inversion failed, or the ABMs catastrophically failed. Additionally, for most categories, we chose the former half of image-text pairs for the partial dataset. We did not cherry pick the images to improve our method’s results. When testing the ABMs, we found some methods had a high number of failure cases on the claimed specialized type.  Finally, we did not include editing types (style and color change) where ABMs required a large search to fine-tune the hyperparameters, since this is infeasible due to some method’s computational complexity and lack of guidelines for searching, if we want to compare the results with our best hyperparameters. In total, 208 images were selected.

For the sub-categories results shown in Tab.2, we further correct the partial dataset from the mentioned issues, and 203 images are selected.

Overall, our experiments are conducted fairly since we select the images that the comparison ABMs perform well on. All editing results and hyperparams will be released with the source code.

We evaluated the CLIP score across the entire image and the LPIPS score for the background region, with results detailed in Tab.~\ref{table:quan}. While our method exhibits slightly better over other ABMs, we do not consider the CLIP score to be an effective metric for evaluating editing quality. This is because, according to human perception, the editing results produced by P2P are generally better than those by PNP.

\subsection{Qualitative results}
\begin{figure}[!ht]
    \centering
    \includegraphics[width=0.97\textwidth]{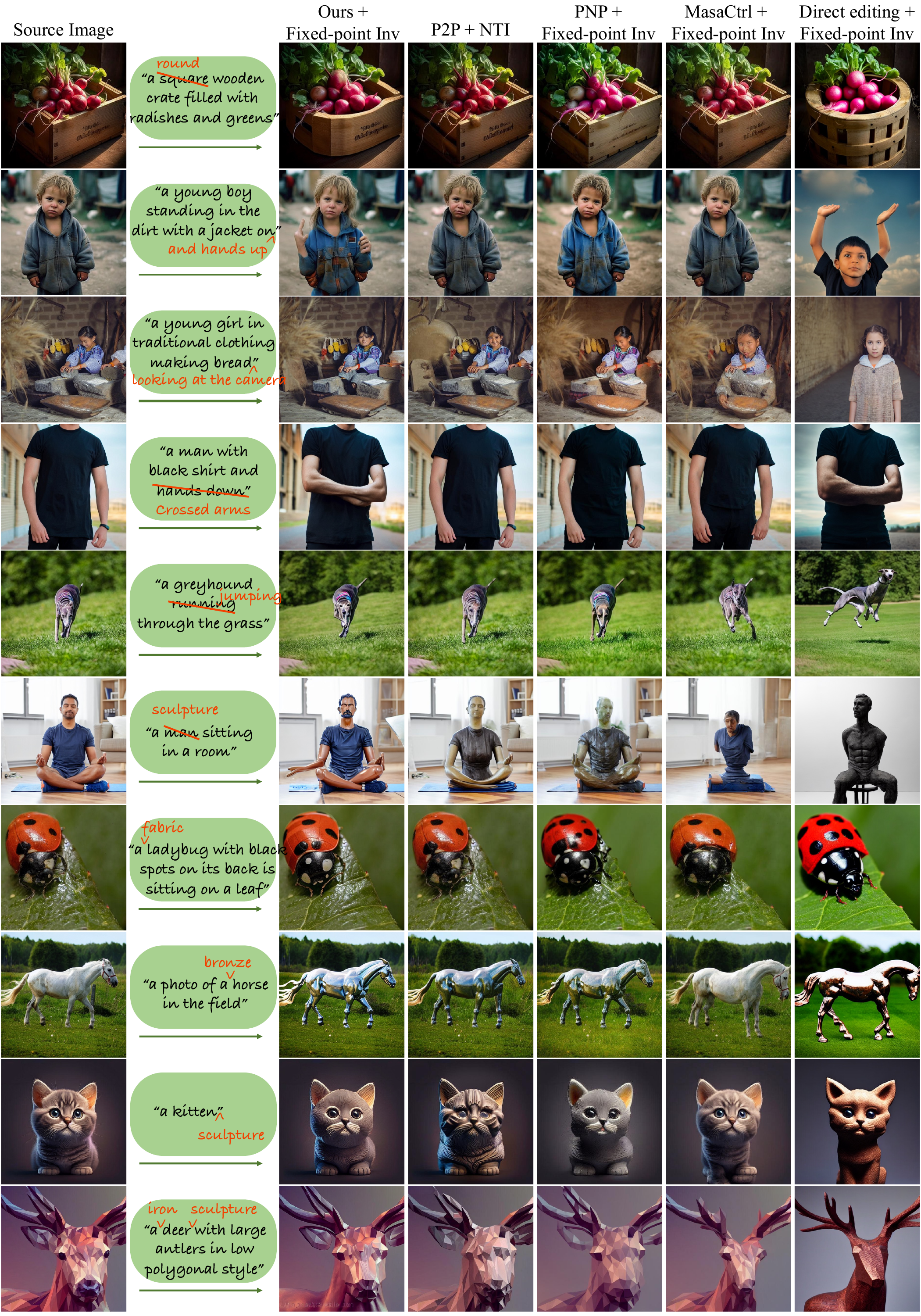}
    \caption{Qualitative comparisons in changing materials, altering poses, and shapes using images from the PIE dataset\cite{ju2023direct}
    . The analysis juxtaposes our approach with 3 typical ABMs: P2P, PNP, and MasaCtrl. Direct editing results with fixed-point inversion are also included as a baseline for benchmarking.}
    \label{fig:qual-0}
\end{figure}

\begin{figure}[!ht]
    \centering
    \includegraphics[width=0.97\textwidth]{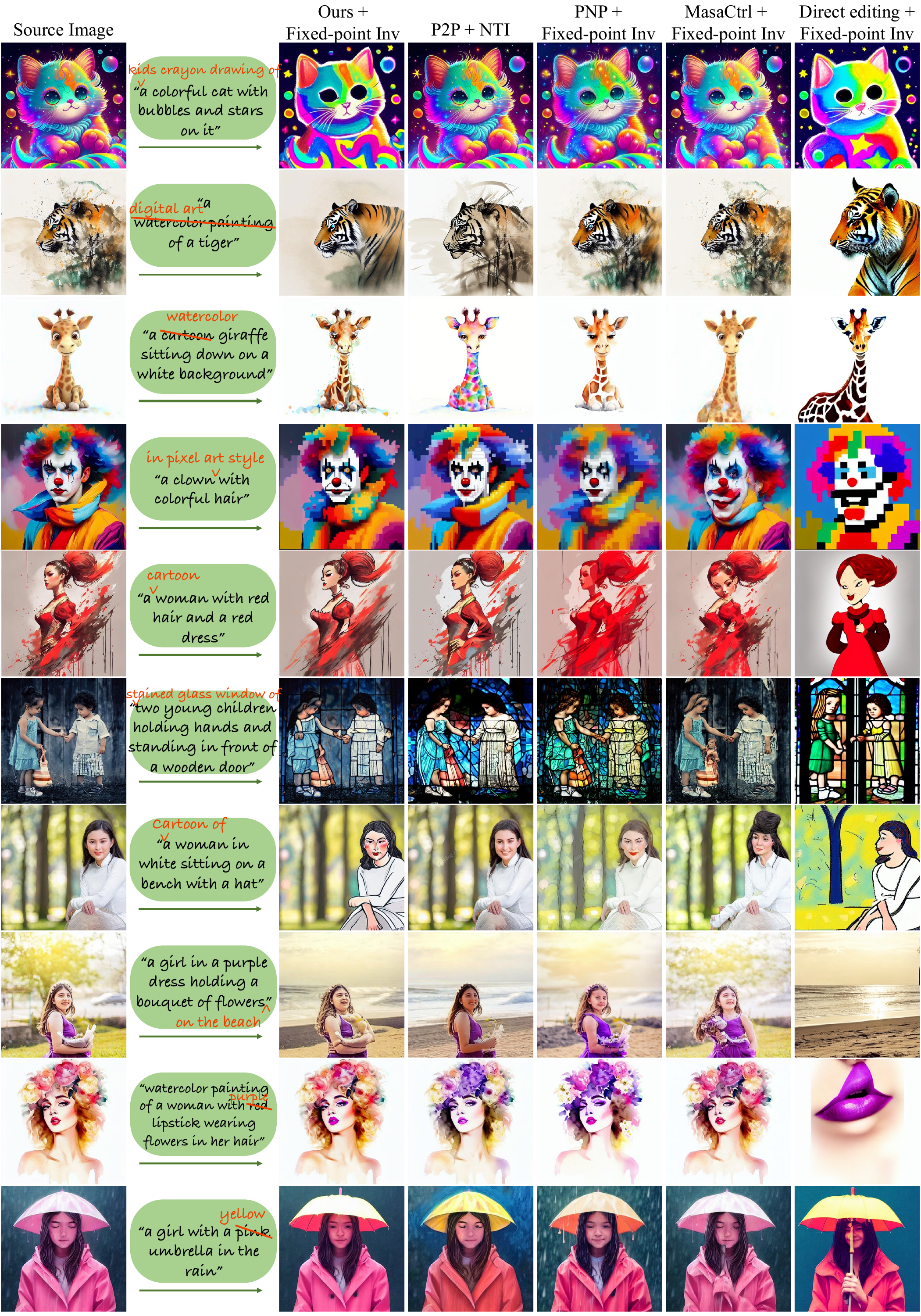}
    \caption{Qualitative comparisons in changing styles, colors, and environment using images from the PIE dataset\cite{ju2023direct}
    . The analysis juxtaposes our approach with 3 typical ABMs: P2P, PNP, and MasaCtrl. Direct editing results with fixed-point inversion are also included as a baseline for benchmarking.}
    \label{fig:qual-1}
\end{figure}

\begin{figure}[!ht]
    \centering
    \includegraphics[width=0.97\textwidth]{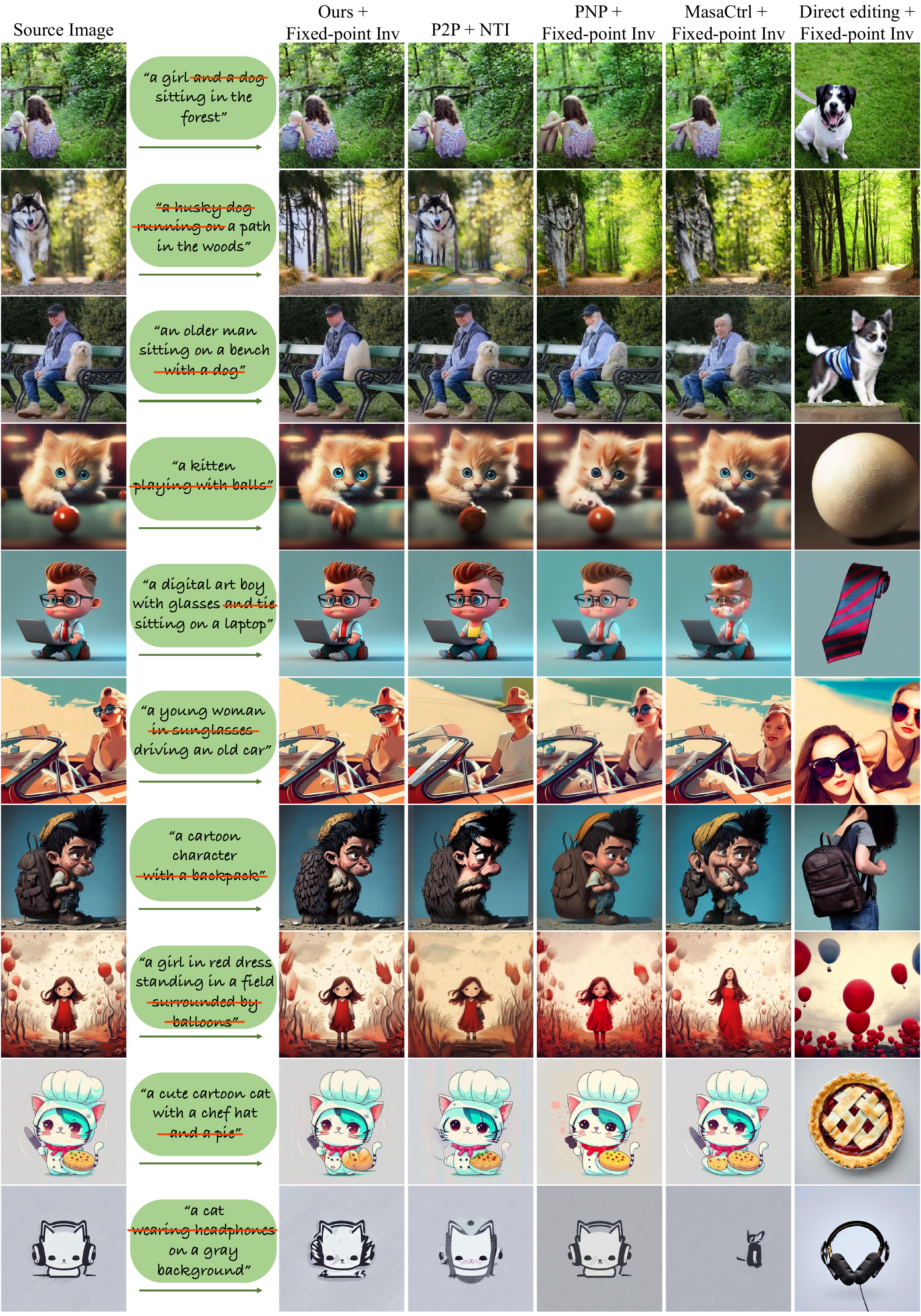}
    \caption{Qualitative comparisons in removing objects using images from the PIE dataset\cite{ju2023direct}
    . The analysis juxtaposes our approach with 3 typical ABMs: P2P, PNP, and MasaCtrl. Direct editing results with fixed-point inversion are also included as a baseline for benchmarking.}
    \label{fig:qual-2}
\end{figure}

\begin{figure}[!ht]
    \centering
    \includegraphics[width=0.97\textwidth]{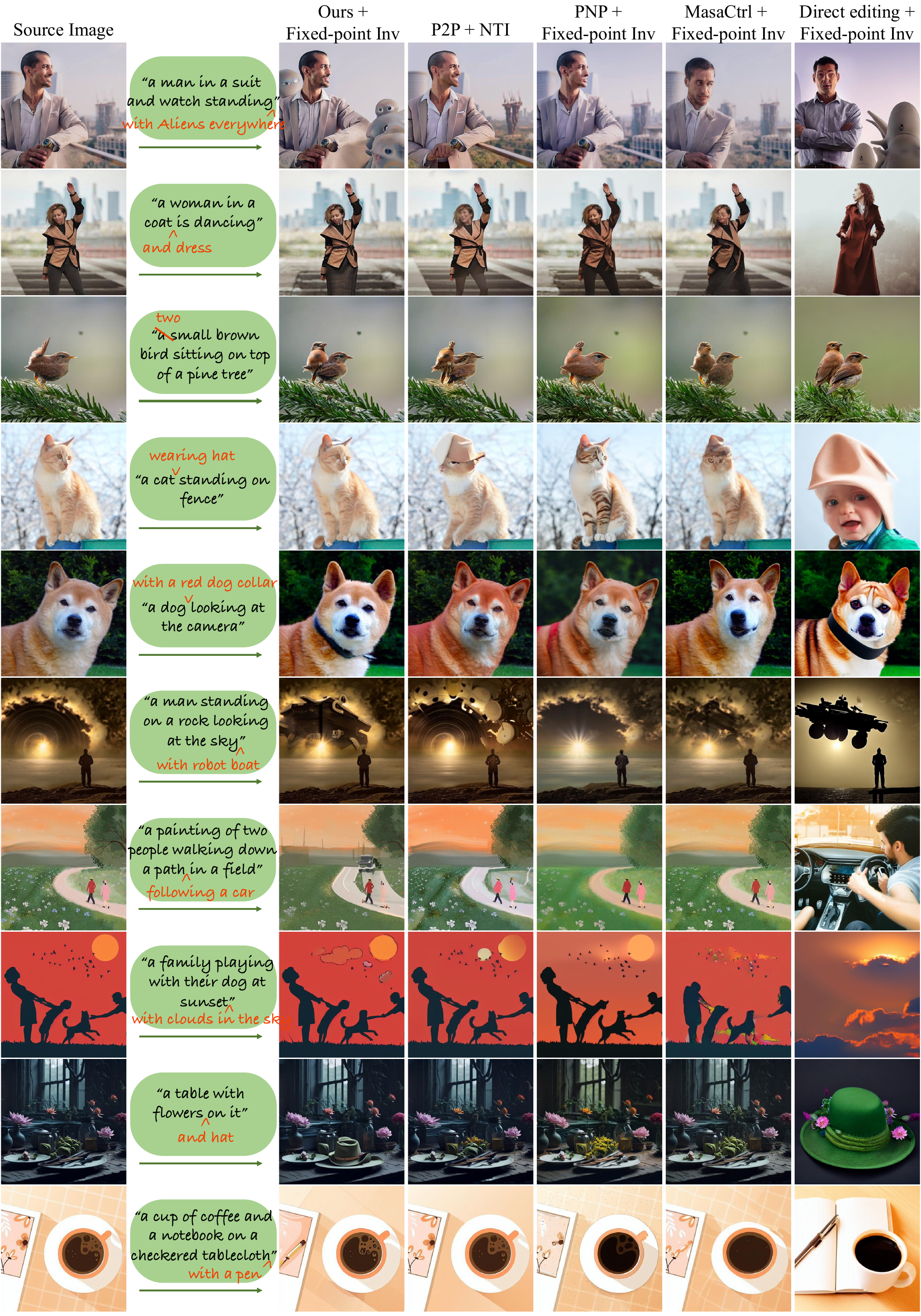}
    \caption{Qualitative comparisons in adding objects using images from the PIE dataset\cite{ju2023direct}
    . The analysis juxtaposes our approach with 3 typical ABMs: P2P, PNP, and MasaCtrl. Direct editing results with fixed-point inversion are also included as a baseline for benchmarking.}
    \label{fig:qual-3}
\end{figure}

\begin{figure}[!ht]
    \centering
    \includegraphics[width=0.97\textwidth]{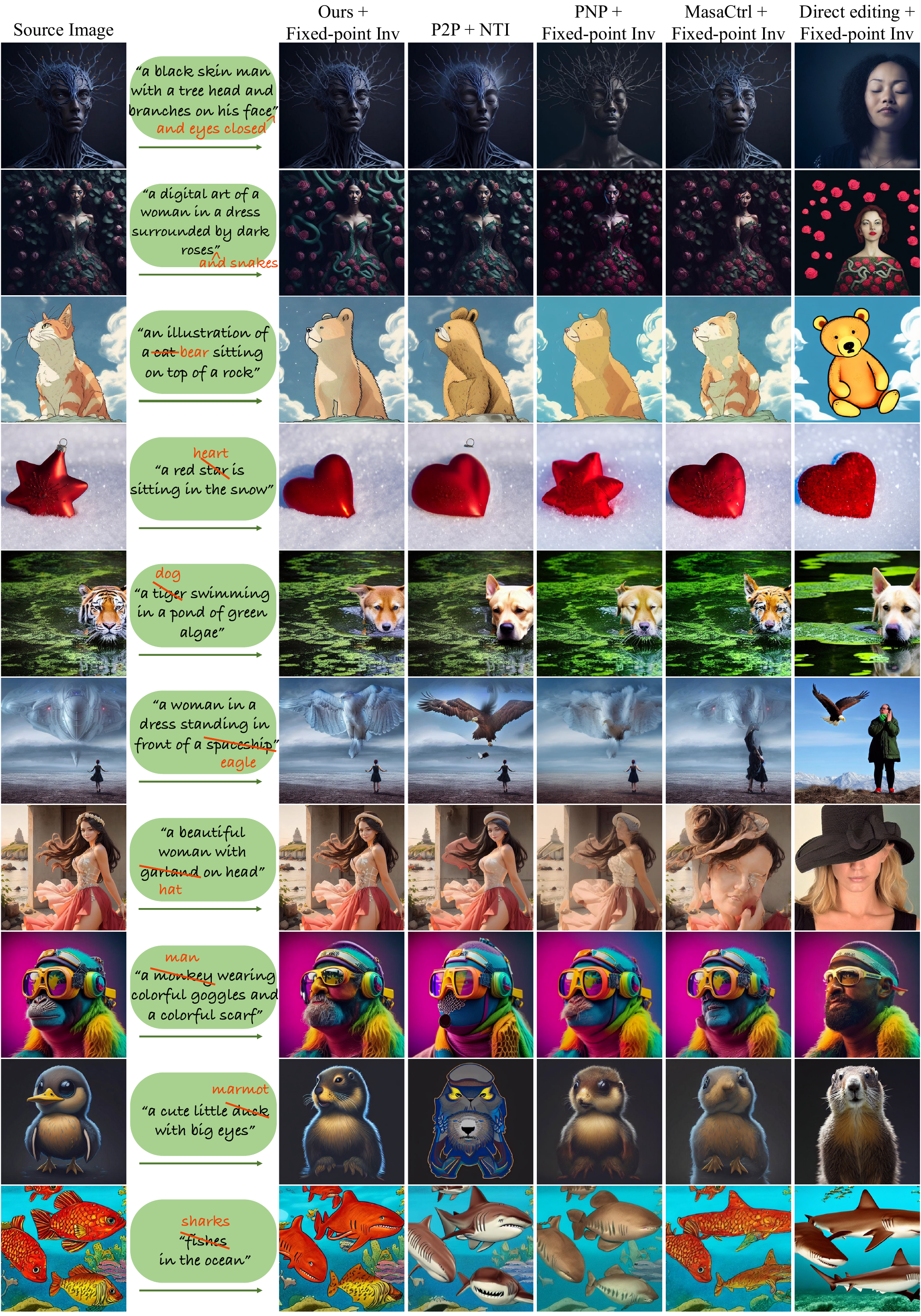}
    \caption{Qualitative comparisons in changing identities, altering shape, and adding objects using images from the PIE dataset\cite{ju2023direct}
    . The analysis juxtaposes our approach with 3 typical ABMs: P2P, PNP, and MasaCtrl. Direct editing results with fixed-point inversion are also included as a baseline for benchmarking.}
    \label{fig:qual-4}
\end{figure}

A wide range of examples across all editing types in the figures attest to the effectiveness of our proposed method. For changes in materials, see Fig.~\ref{fig:qual-0}; for changing styles, colors, and environments, see Fig.~\ref{fig:qual-1}; for removing objects, see Fig.~\ref{fig:qual-2}; for adding objects, see Fig.~\ref{fig:qual-3}; for changing in identities, see Fig.~\ref{fig:qual-4}.

\end{document}